\documentclass[letterpaper, 10 pt, conference]{ieeeconf}  

\IEEEoverridecommandlockouts                              

\usepackage{graphicx}
\usepackage{subfigure}
\usepackage{verbatim}
\usepackage{amssymb}
\usepackage{amstext}
\usepackage{amsmath}
\usepackage{url}
\usepackage[usenames,dvipsnames]{color}
\usepackage[normalem]{ulem}
\graphicspath{{figures/}}

\usepackage{listings}   
\definecolor{dkgreen}{rgb}{0,0.6,0}
\definecolor{gray}{rgb}{0.5,0.5,0.5}
\definecolor{mauve}{rgb}{0.58,0,0.82}
\definecolor{armygreen}{rgb}{0.29, 0.33, 0.13}
\definecolor{battleshipgrey}{rgb}{0.52, 0.52, 0.51}
\definecolor{airforceblue}{rgb}{0.36, 0.54, 0.66}
\definecolor{babyblueeyes}{rgb}{0.63, 0.79, 0.95}
\definecolor{babypink}{rgb}{0.96, 0.76, 0.76}
\definecolor{cornellred}{rgb}{0.7, 0.11, 0.11}

\definecolor{applegreen}{rgb}{0.55, 0.71, 0.0}
\definecolor{darkorange}{rgb}{1.0, 0.55, 0.0}

\newcommand{\Figure}[0]{Fig.~}

\newcommand{\Section}[0]{Section~}

\title{\LARGE \bf
Towards Error Handling in a DSL for Robot Assembly Tasks* 
}

\author{
Johan {S. Laursen},
\and Jacob {P. Buch},
\and Lars {C. S{\o}rensen},
\and Dirk {Kraft},
\and Henrik {G. Petersen}
\and Lars-Peter {Ellekilde},
\and and Ulrik {P. Schultz}$^{1}$
\thanks{*This work was supported by The Danish Council for Strategic Research through the CARMEN project.}
\thanks{$^{1}$The M{\ae}rsk Mc-Kinney M{\o}ller Institute, University of Southern Denmark, Odense, Denmark\
        {\tt\small \{josl, jpb, lcs, kraft, hgp, lpe, ups\}@mmmi.sdu.dk}}%
}

\begin{document}

\maketitle
\thispagestyle{empty}
\pagestyle{empty}


\begin{abstract}
This work-in-progress paper presents our work with a domain specific language (DSL) for tackling the issue of programming robots for small-sized batch production. We observe that as the complexity of assembly increases so does the likelihood of errors, and these errors need to be addressed.  Nevertheless, it is essential that programming and setting up the assembly remains fast, allows quick changeovers, easy adjustments and reconfigurations.

In this paper we present an initial design and implementation of extending an existing DSL for assembly operations with error specification, error handling and advanced move commands incorporating error tolerance. The DSL is used as part of a framework that aims at tackling uncertainties through a probabilistic approach.
\end{abstract}

\section{INTRODUCTION}
Bringing robotics to small-sized production is highly important~\cite{Martin2005}. The traditional approach within automation has taken advantage of robots high repeatability and ensures deterministic behavior of systems through highly customized components and fixation. This approach is however not applicable and relevant for many small-sized batch productions where automation systems have to be able to cope with high variance, high flexibility and be capable of handling inaccuracies and pose-uncertainties while remaining easy and intuitive to use. 

To overcome these issues, we are working on a software framework and concept that use a probabilistic approach for active handling of uncertainties. The framework is centered on an action library where actions are parameterized. Simulation is used to facilitate learning of uncertainty-tolerant actions through an optimal choice of parameters. 

More complex assembly tasks also results in more complex frameworks, programming, assembly operations and algorithms. High complexity also means additional places for things to go wrong and hence more errors may occur.

We take an approach where it is assumed that errors are inevitable and will appear at some point during the assembly. Therefore, instead of avoiding errors, the errors are to be managed and rectified. 
This is achieved by integrating error specification and management into the domain-specific language (DSL) used to program assembly task through the framework. Moreover, we define an error-aware move instruction suitable for operation sequences where an error is likely to occur during an assembly process. These instructions employ simple sensor readings and based on user-specified conditions informs the system of error occurrences.
The DSL builds and expands upon an already existing DSL for programming assembly tasks through the framework~\cite{CARMEN2014}.
The paper is structured with \Section\ref{sec:system} explaining the overall concept and software framework, \Section\ref{sec:dsl} presents the DSL improvements, the BNF and a syntax example. Next \Section\ref{sec:experiment} details implementation and testing, and \Section\ref{sec:future} describes future work and how we intend to make error recovery autonomous through reverse execution. The paper is concluded in \Section\ref{sec:conclusion}.


\section{SYSTEM CONCEPT AND ARCHITECTURE}
\label{sec:system}
 The main focus of this paper is to demonstrate our ideas and concepts regarding the inclusion of error specification and advanced error-aware move functions into a DSL for programming robots. The DSL is part of a probabilistic software framework, which is designed for quick derivation of assembly solutions. Today a significant amount of work goes into ensuring deterministic behavior of objects when preparing tasks for automation. Whether this is done by creating customized fixtures, grippers and external equipment or by integrating different and advanced sensors, this is a time consuming process.

The framework is created to reduce the need of deterministic behavior through software. Instead of having to eliminate positional uncertainties and minor variation of objects the variations are to be handled and managed. In the framework we do this by creating actions and robot movements that are able to cope with a high degree of variation in the objects position; The idea and concept is discussed in greater detail in \cite{CARMEN2014}. However we want to further extend this idea of handling uncertainties to the way we program assembly tasks. We are therefore looking at how to create a language which allows use of the probabilistic actions while also incorporating its own constructs for handling uncertainties. As a start, we do this by bringing error management and handling into our DSL. This approach also means that the language is more relevant for small-sized production rather than large scale assembly where property like constant tick-time and throughput are key properties, since this implies a high repeatability and predictability of the systems.

%

Three topics are of special interest in our development of the software framework: (1) Applying mathematical models and simulation techniques for fast and realistic dynamic simulation to handle the number of experiments required to learn action parameters; (2) Formalizing probabilistic actions such that they can handle inaccuracies; (3) Interfacing to these actions as well as using and employing them through a DSL designed to solve the assembly tasks.

\subsection{Description of the Software Framework}
\begin{figure}[t] 
  \centering
  \includegraphics[width = 0.45\textwidth]{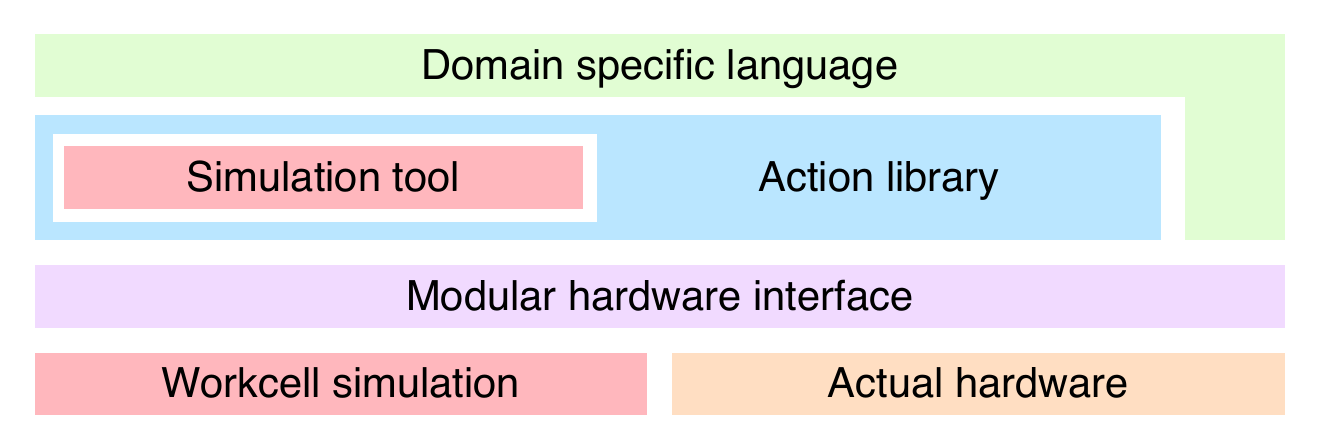}
  \caption{The four-layered architecture of the software framework.}
  \label{fig:Architecture}
\end{figure}

A four-layer system architecture is used as the foundation of the software framework for the concept. The architecture 
takes inspiration from the SoftRobot~\cite{Angerer2013} project.
 The architecture is illustrated in Figure~\ref{fig:Architecture} and explained below.

\noindent\textbf{Workcell Simulation/Actual Hardware:}
On the lowest level is the actual hardware. The hardware is chosen based on the required needs. A simulation of the workcell is made, enabling a complete test of the system in both simulation and real world, as required by the probabilistic actions.

\noindent\textbf{Modular Hardware Interface:} 
To ensure that the framework can be adapted and converted to new hardware without too much effort, the hardware is interfaced through standardized interfaces which can be combined in a modular fashion. 
The interfaces include robot commands as well as digital and analogue I/O, grippers, sensors and so forth; 
they provide a basis for the hardware-near actions

\noindent\textbf{Action Library (Simulation Tool):} 
The third layer provides structuring of more complex actions.
Here the probabilistic aspect is explicit 
and uncertainties are modeled and accounted for rather than avoided. Actions from the action library are modeled through sets of parameters, where a given set instantiates a corresponding executable action.  
Simulation is used to facilitate the learning of parameters making the action tolerant to pose uncertainties. 
All actions therefore contain a precondition, where the state of relevant parts is specified. Sampling and learning is done using simulation, instead of performing the corresponding real experiments, as it significantly reduces costs and time.
In order to simplify the creation and implementation of new actions, a generic interface is used to communicate between actions and simulation. 

\noindent\textbf{Domain Specific Language:} The top layer and interface to the framework comes in the form of a DSL. As in \cite{Thomas2013} the DSL embarks on programming with actions and skills. 
While hardware and concurrent simulation is interfaced from the actions within the library they can also be interfaced through the DSL. 
This allows the user to program the hardware to performed complex tasks through a combination of more abstract and high-level actions along with more hardware-near and normal robot operation instructions. The aim is to have a system which is programmed in an easy and intuitive way.

We will now present our current effort towards further embracing the probabilistic aspects of the concept and framework to an even higher degree in the DSL. We aim to do this by integrating user-specified errors and error handling, and by allowing more advanced error-aware move commands to be specified through the DSL. These advanced move commands take advantage of sensor readings to evaluate if the move was successful and can signal the specified errors to the system such that appropriate methods for solving the problem can be taken. 
The error handling takes inspiration form the try/catch-construct and the approach taken in~\cite{Simmons98}. When an error is thrown the controller executes a specified recovery sequences. Our approach however differs in the amount of information attached to an error: Information such as how to proceed afterwards and the severity of the error is included. 


\section{THE DOMAIN SPECIFIC LANGUAGE}
\label{sec:dsl}
A key part of the software framework is the DSL which serves as the primary programming interface for the user. 
The DSL is developed to reduce complexity and allow easy programming of assembly tasks using the features of the framework. Through the DSL individual actions and robot commands can be orchestrated and combined. 
As in \cite{Henrik2010} the DSL is approached and styled around a traditional robot programming language and the flexibility of the DSL allows us to modify and extend the language with new features. 
The software framework is still subject to development and large-scale modifications, and therefore the DSL is also currently implemented as an internal DSL in C$++$ to ensure it remains tightly integrated and up-to-date with the underlying framework.
Moreover, keeping the DSL as an internal C$++$ DSL gives the developers working with the framework, and who are already comfortable working in C$++$ and their C$++$-environments, the ability to experiment with the DSL and associated features more easily.
The original DSL was presented in earlier work~\cite{CARMEN2014}, this paper presents an extension of the DSL for error handling; for readability the DSL is however presented as a whole in this section. 

The DSL allows the user to interface the robot and platform on different abstraction levels through sets of instructions grouped into sequences. These instructions range from hardware-near robot instructions and I/O-commands to abstract actions. The DSL also allows the user to specify complex move instructions to the robot. Here the user specifies an intended target along with a set of conditions specifying the requirements for the move to be successful. Depending on the outcome, a user-specified error is thrown. This error is specified within the DSL, and the controller responsible for the execution of sequences and instructions manages and resolves these errors as they are thrown. The errors include parameters to identify the error, the action needed to be taken if the error is encountered, how fast the response time needs to be, and how to continue operation after the error has been resolved.


\subsection{The abstract syntax BNF}
Omitting syntactic noise due to the use of an internal DSL, the BNF of the \emph{abstract syntax} of our DSL is shown in Fig.~\ref{fig:syste:bnf}. 
Above the line break correspond to the original language (presented in \cite{CARMEN2014}) while the rest concerns the additions of errors and advanced move commands (this paper).

The DSL allows I/O-operations to be specified \emph{(rule IO)} 
in terms of primitives and wait commands to allow greater control of external equipment. Joint configurations for simple move commands can be specified \emph{(rule JointConf)} along with meta information associated with the assembly items and used by the actions \emph{(rule Item)}. 
User-specified errors are defined and provide information regarding how the system should respond to the error \emph{(rule Error)}. The advanced error-aware move commands are defined from \emph{(rule Amove)} and provide information regarding how to move, the success criteria of the move, and how to respond if the move succeed or fails. 
From this information a sequence of operations can be defined \emph{(rule Sequence)}. These sequences consist of simple and advanced move commands, actions from the action library, I/O-operations, control commands and previously defined sequences. All together this creates a building block structure where actions can be combined, modeled and reused.

\newcommand{\xor}{$|$ }
\newcommand{\xterm}[1]{\underline{#1}}
\newcommand{\xnterm}[1]{\emph{#1}}
\newcommand{\xmany}{$\ast$ }
\newcommand{\xone}{$+$ }
\newcommand{\xnone}{$?$ }
\newcommand{\xeq}{$\::=$}
\newcommand{\xsub}[1]{\mbox{\tiny\sf #1}}
\def\arraystretch{1.1}%
\setlength{\tabcolsep}{3pt}
\begin {figure}[t]
\scriptsize
\centering
\begin{tabular}{lcl}
\xnterm{P} 			&\xeq& (	\xnterm{Item} 		\xor{}
							 	\xnterm{IO} 		\xor{}
							   	\xnterm{JointConf} 	\xor{}
							   	\xnterm{Sequence} 	\xor{}
							   	\xnterm{Error} 		\xor{} 
							   	\xnterm{AdvMove} ) 	\xmany \\
\xnterm{Item} 		&\xeq& \xterm{item} $N_{\xsub{item}}$ ( \xterm{keyframe} $N$ \xnterm{coordinate}\xmany ) \\
\xnterm{IO} 		&\xeq& \xterm{IOoperation} $N_{\xsub{io}}$ \xnterm{Primitive}\xmany \\
\xnterm{JointConf} 	&\xeq& \xterm{JointConfiguration} $N_{\xsub{joint}}$ \xnterm{vector} \\
\xnterm{Sequence} 	&\xeq& \xterm{sequence} $N_{\xsub{seq}}$ \xnterm{Action}\xmany \\
\xnterm{Primitive} 	&\xeq& \xterm{setLow} \xor{} \xterm{bit} $n$ \xor{} \xterm{sleep} $n$ \xor{} \ldots \\
\xnterm{Action} 	&\xeq& \xterm{move} $N_{\xsub{joint}}$ 
								\xor{} \xterm{call} $N_{\xsub{action}}$($N_{\xsub{item}}$\xmany) 
								\xor{} \xterm{io} $N_{\xsub{io}}$ \xor{} \xterm{wait} $n$ \xor{} \ldots \\[3ex]

\xnterm{Error}		&\xeq& \xterm{error} $N_{\xsub{error}} $  ( \xterm{recoverySequence} $N_{\xsub{seq}}$ ) \xnone 
							 					   				  \xnterm{Etime}\xnone  \xnterm{Eoption}\xnone\\
\xnterm{Etime} 		&\xeq& \xterm{respondAfter} ( currentAction \xor currentSequence \xor ... ) \\
\xnterm{Eoption}	&\xeq& \xterm{returnTo} ( action \xor sequence \xor restartProgram \xor ... ) \\
\xnterm{AdvMove}	&\xeq& \xterm{move} $N_{\xsub{move}} $ \xnterm{Acondition} \xnone \xnterm{Aspeci} 
															 \xnterm{Aeval}\xnone \\
\xnterm{Aspeci}		&\xeq& \xterm{specification} \xnterm{Amove} ( \xterm{stop if} \xnterm{Aquery} ) \xnone 
													   \xnterm{Asetting} \xnone \\
\xnterm{Aeval}		&\xeq& \xterm{evaluation} \xnterm{Aquery}\xone (\xterm{onSuccess:}\xnterm{Abehave}\xmany)\xnone (\xterm{onFail:}\xnterm{Abehave}\xmany)\xone \\
\xnterm{Amove}		&\xeq&	\xterm{distance(} $n$ \xterm{,} \xnterm{direction} \xterm{,} \xnterm{frame} \xterm{)}\\
\xnterm{Aquery}		&\xeq&	\xterm(forcesExceeds($n$)  \xor distanceCoverdExceeds($n$) 
														   \xor ... ) \xone \\
\xnterm{Asetting}	&\xeq&	\xterm{settings} \xnterm{speed}\\
\xnterm{Abehave}	&\xeq&	( returnToInitPos  \xor repeatWperturbation($n$) 
												   \xor throwError($N_{\xsub{error}} $) )	\\
\xnterm{direction}	&\xeq&	\xterm{direction} ( forward \xor backwards \xor left \xor ... \xor x \xor y \xor z)\\
\xnterm{frame}		&\xeq&	\xterm{frame} ( tcp \xor toolmount \xor base )\\
\xnterm{speed}		&\xeq&	\xterm{speed} ( very fast \xor fast \xor ... \xor slow \xor very slow )\\
\multicolumn{3}{l}{$N\in$ names, $n\in\mathbb{R}$, \xnterm{coordinate} specifies coordinates, \xnterm{vector}$\in\mathbb{R}^n$}
\end{tabular}
\caption{BNF of the abstract syntax for the internal DSL.}
\label{fig:syste:bnf}
\end{figure}
\def\arraystretch{1.0}%

\subsection{Usages example code}
\label{sec:example}
We now use a small example to illustrate the DSL. The task is to perform a peg-in-hole assembly operation with a tight fit. First, platform-specific information regarding I/O-operations and joint configurations are declared (see \Figure \ref{dslcode:1}).

\begin{figure}[t] 
\centering
\begin{lstlisting}[frame=]
IOoperations().
  manipulation("gripper_open").
    setLow().bit(0).sleep(0.5).
  manipulation("gripper_close").
    setHigh().bit(0).sleep(0.5);

JointConfiguration
  startPosition    = {3.425, -1.0...},
  handlePosition   = {3.379, -1.2...};
\end{lstlisting}
\caption{Code example: Declaration of platform information. }
\label{dslcode:1}
\end{figure}

Next, a user-specified error stating that the peg was not inserted is defined along with a sequence describing how the system should manage the error (see \Figure \ref{dslcode:2}).

\begin{figure}[t] 
\centering
\begin{lstlisting}[frame=]
sequence("peg_in_hole_recovery").
  move().to(startPosition, handlePosition).
  io("gripper_open").
  ...
  move().to(startPosition);

Error( "peg_not_inserted" )
  .recovery_using_sequence( "peg_in_hole_recovery" )
  .recovery_starts_after( currentAction )
  .postrecovery_behaviour( returnToSequence );
\end{lstlisting}
\caption{Code example: User-defined errors and management sequence.}
\label{dslcode:2}
\end{figure}

Finally, an advanced error-aware move is defined. Here, the peg is moved forward, but the robot is instructed to stop if forces exceed a max force. This occurs if the peg and hole is unaligned. If the move succeeds the peg is moved back to its initial position, but if unsuccessful it moves back to its initial position, adds a random perturbation to its position, and attempts the move again. This is attempted three times before signaling the error (defined in \Figure\ref{dslcode:2}). A controller handles this error signal by activating the error management sequence. The advanced move is used as a part of a sequence, which ensures the gripper is open and moves to the initial position before attempting the advanced move (see \Figure \ref{dslcode:3}).

\begin{figure}[t] 
\centering
\begin{lstlisting}[frame=]
advanced_move( "insert_peg" ).
specifications()
      .distance(0.30, direction::forward, frame::tcp)
      .stop_ifForcesExceed(5)
      .setting(speed::slow).
evaluation()
      .distanceCovered(crocodile::moreThen, 0.20).
behaviour_on_success()
      .returnToInitialPosition().
behaviour_on_failure()
      .returnToInitialPosition()
      .repeatMoveWithPerturbations(3)
      .throwError( "peg_not_inserted" );

sequence("reliability_test").
    io("gripper_close").
    move().to(startPosition).
    AdvMove("insert_peg").
    move().to(startPosition);
\end{lstlisting}
\caption{Code example: Advanced move and main sequence.}
\label{dslcode:3}
\end{figure}

\section{IMPLEMENTATION AND EXPERIMENTS}
\label{sec:experiment}
\begin{figure}[t] 
  \centering
  \includegraphics[width = 0.2\textwidth]{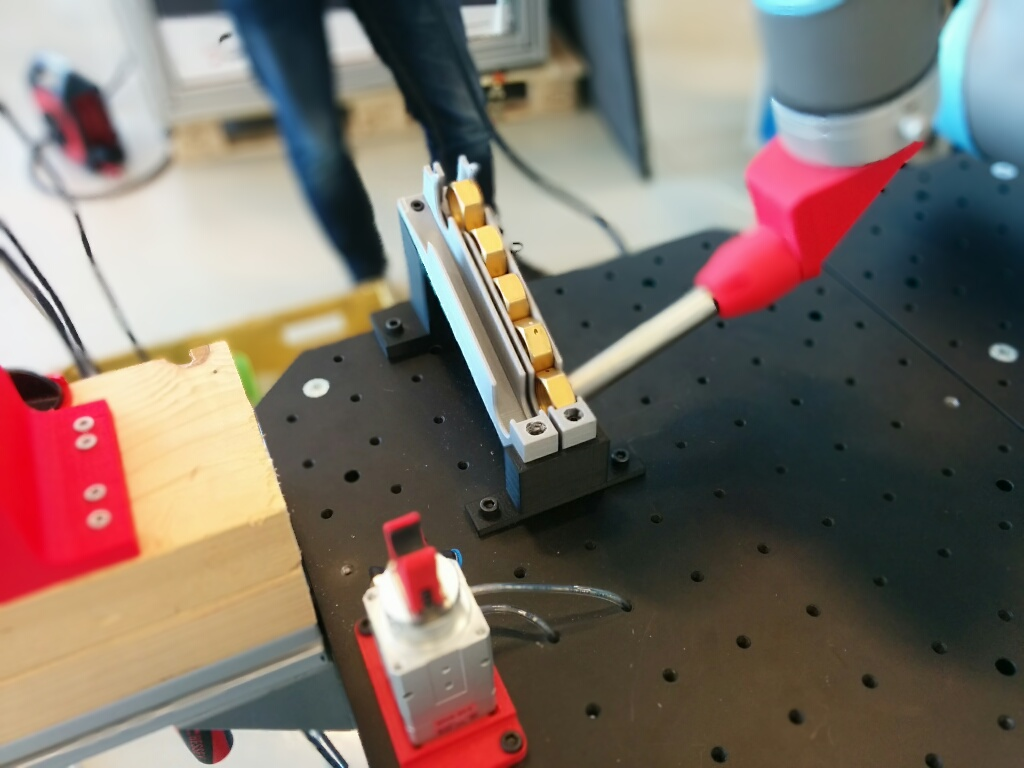}
  \includegraphics[width = 0.2\textwidth]{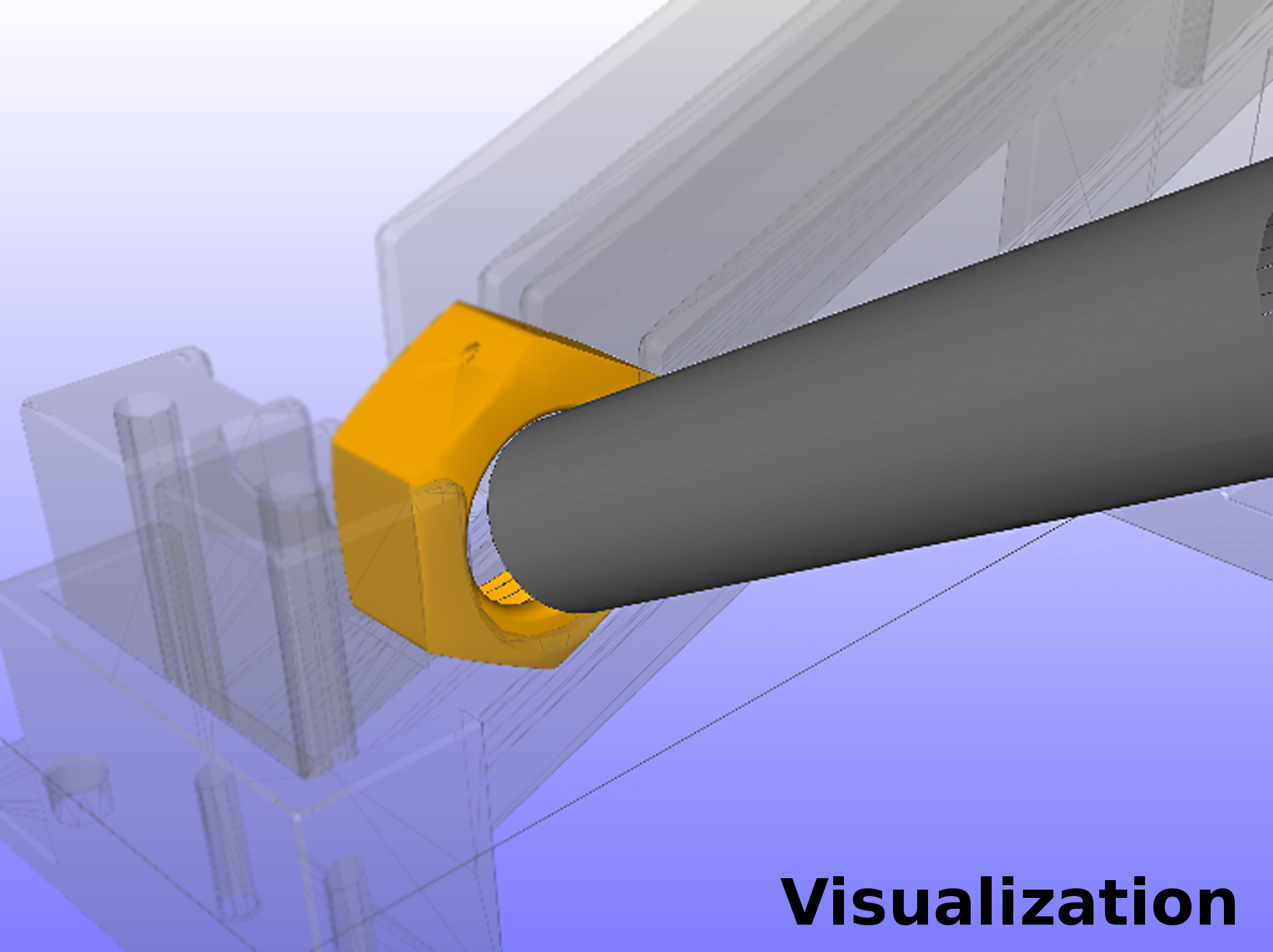}
  \caption{The set-ups used to test the DSL.  User-specified errors and advanced move commands are only tested on the kinematic simulation.}
  \label{fig:experiment}
\end{figure}

Those parts of the DSL that concern user-specified errors and advanced error-aware moves have so far only been tested using a kinematic simulation, whereas other parts such as actions, I/O-operations and hardware-near robot commands has been successfully deployed through the framework to a Universal Robots UR5 robot. \Figure \ref{fig:experiment} shows the real set-up and the kinematic simulation workspace.

The DSL employs a standard model-based approach where an internal meta-model is instantiated and subsequently interpreted. This allows us to build all components separately before they are combined. A controller class is responsible for the execution of sequences. When an error is thrown the controller class is notified and can extract information regarding when and how to handle the error.

Instead of using the pipeline to connect to a robot, the hardware interface was replaced with a kinematic simulation 
from the RobWork framework~\cite{Ellekilde2010}. 
Through this kinematic simulation it is possible to visualize the work cell and the movements of the robot. If the robot collides with its surroundings it returns a predefined force as its force/torque readings. The simulation thus provides no data which the UR5 robot is not able to provide in real life and so the program should be executable on the robot as soon as the communication protocol from the modular hardware interface is extended to include the robots force/torque and servo commands. The real force/torque measurements are however subject to significant noise and we expect that a running-average filter will need to be implemented.


Small examples such as the code explained in \Section \ref{sec:example} have been successfully executed on the kinematic simulation. From this it can be seen that the functionality and logic of the error management works as intended.

Through this approach of combining the use of advanced error-aware move functionality with error handling we have been able to solve small tasks through a trial-and-error approach. We find it natural 
and simple to apply this trial-and-error approach to solve tasks with the DSL rather than by programming the robot though traditional means. Many complex assembly tasks sensible to uncertainties might take advantage of this approach rather than having to ensure deterministic behavior through customized components and fixation. How the branching and forking of sequences due to error handling affects the program on larger tasks is subject to future experimentation with the real robot.

\section{FUTURE WORK: ERROR RECOVERY THROUGH REVERSE EXECUTION OF SEQUENCES}
\label{sec:future}
The current work presents a system in which error conditions and recovery sequences are explicitly provided by the user and straightforward executed by the controller. In the future we aim to make this more autonomous with automatic error detection based on a probabilistic framework comparing current and past executions. Rather than having explicit static recovery sequences we would give the controller the ability to fix certain errors itself through reverse execution of sequences and backtracking through the program. Upon resuming forward execution many errors will hopefully fix themselves, due to the probabilistic nature of many uncertainties and sensor readings. Otherwise a more active approach could be taken, where small perturbations to the robots movements are added or parameters changed to make the execution more conservative.

Most commercial robot controllers already allow the user to step through programs in reverse, however this is mainly as a programming and debugging tool. When reversing through a program, these controllers employ an approach where they only reverse the sequence of commands. Each command is executed with no regard to using the obverse or reverse counterpart of the command. The controllers also provide no logic for different sequencing structures such as branching and forking, these are simply skipped or can not be reversed. What we propose is to apply the concept of reverse execution to automatic error recovery such that much of the reverse execution process is either automated or specified through the program. This eliminates the need for an operator to constantly determine whether a step actually is reversible.  

\subsection{Reversibility of processes and operations}
There are two significant problems which need to be addressed in this reverse execution of assembly, one of which is that not all parts of an assembly process are reversible. An object might have been deformed, two objects might have been clicked together or it might not be possible to return an object grasped from a feeder to the position from where it was grasped.
The second problem arises from the fact that the reverse counterparts to language primitives and robot commands may change depending on the context and abstraction of which they are applied. In a simple case the reverse of switching an I/O-port to on could be to switch that very same I/O port off. However if the I/O-port was used to enable equipment before an operation, such as enabling power to a gripper before opening it, reverse execution would be more complex. 

We hope to overcome these problems by providing and combining constructs regarding reversibility, ordering and sequencing structures. 

Reversibility is split into three categories: Always reversible, reversible after forward execution and never reversible. Using these definitions we can flag different parts of the assembly process accordingly if the process is not reversible. We also use the categories to assign default obverse and reverse counterparts to commands. Commands such as \textit{toggle I/O} are easy to categorize but many commands have varying degrees of ambiguity making it more difficult. 
A move instruction could upon forward execution move the robot from a random initial start position to the specified destination. 
Upon reverse would the robot then go back to its random initial position or only return to first position specified in the program? The two options would mean that the move was respectively reversible only after forward execution or always reversible. 
Another example of a command which contains this ambiguity is a force sensitive move command. Would the reverse of an instruction to apply a force in a direction be to apply the force in opposite direction or just move the robot back to its initial position? 
Commands are not necessarily limited to one category, but they are provided with a default category to avoid clutter in the programming language.  
When move commands employ the concept of only being reversible after forward execution we call them kinematic reversible. The concept of kinematic reversibility also gives a clear separation and abstraction between the programming language and the probabilistic and high-level actions in the framework, where kinematic reversible is the default option. 

\subsection{Design of a reversible language}
We are also intent on providing constructs and structures for sequencing and ordering of command execution. Here we want to organize commands in different patterns that match their reversibility. 
Some of the most simple and widespread patterns within robotic assembly automation is sequences and hierarchical structures. Reversing a sequence is fairly simple but there are more options regarding how to reverse hierarchical structures. 
We are also looking at how to incorporate more advanced structures such as branching and forking. Taking inspiration from reversible computing different approaches can be taken such as simply skipping branches or having both pre and post conditions on statements \cite{yokoyama2008principles}. However robot programs are often smaller than ordinary computer programs and therefore we also explore approaches not viable for computer programs. Recoding all branch decision is for instance entirely feasible in robot applications whereas the amount of data needed to generate and store this in ordinary programs would be tremendous.

We believe that by combining the constructs and categories of reversibility with the knowledge of sequencing structures and ordering it is possible to create programs which are capable of both forward and backwards execution. From the default categories and underlying sequencing model most of the reverse execution can hopefully be automated from the forward execution script.
We do however intent to implementing additional look-ahead features in the underlying execution model. These will help ensure that lasting settings and instructions such as speed is set as intended.


Through the domain-specific advantages of the DSL we also want to incorporate an instruction set from where it can be specified if the reverse execution of the assembly sequence or instruction differentiates itself from the otherwise straight forward and default approach to reversing the program. These instructions could use the constructs to flag sequence parts as non-reversible, designate program points beyond which reversal is not possible or indicate sequence parts to skip for reverse execution. Similarly, changes to parameters, default reverse options or modified execution orders could also be specified.  
Nevertheless, it is our intention that the instruction set be kept simple and non-intrusive, such that it does not divert the users attention from the main tasks and introduce clutter the program, thereby making reading and debugging significant more difficult. The instructions should therefore only be required 
when the default behavior is not appropriate; Experiments regarding instructions usefulness and complexity will therefore be conducted.

If errors are to be solved using reverse execution, it naturally poses the question of when to resume forward execution again. Here, we intend to experiment with different algorithms and logic. An approach could be to reverse further and further back before resuming forward execution each time the same error is met. Learning and memory could also be exploited to make the system aware of how far to reverse before resuming execution.

\section{CONCLUSIONS}
\label{sec:conclusion}
This work-in-progress paper presented our work with a framework for handling automated complex assembly operations in small-size productions through a probabilistic approach. Part of the framework consists of a DSL and the paper focused on how we have incorporated user-defined errors and handling into the DSL. We also explored advanced move commands to make it self-evident for user to attempt to solve assembly tasks through a trial and error approach. Furthermore we have proposed to use reverse execution as a tool for error handling and discussed the associated challenges.

\addtolength{\textheight}{-12cm}


\bibliographystyle{IEEEtran}
\bibliography{carmenref}

\begin{thebibliography}{1}
\providecommand{\url}[1]{#1}
\csname url@samestyle\endcsname
\providecommand{\newblock}{\relax}
\providecommand{\bibinfo}[2]{#2}
\providecommand{\BIBentrySTDinterwordspacing}{\spaceskip=0pt\relax}
\providecommand{\BIBentryALTinterwordstretchfactor}{4}
\providecommand{\BIBentryALTinterwordspacing}{\spaceskip=\fontdimen2\font plus
\BIBentryALTinterwordstretchfactor\fontdimen3\font minus
  \fontdimen4\font\relax}
\providecommand{\BIBforeignlanguage}[2]{{%
\expandafter\ifx\csname l@#1\endcsname\relax
\typeout{** WARNING: IEEEtran.bst: No hyphenation pattern has been}%
\typeout{** loaded for the language `#1'. Using the pattern for}%
\typeout{** the default language instead.}%
\else
\language=\csname l@#1\endcsname
\fi
#2}}
\providecommand{\BIBdecl}{\relax}
\BIBdecl

\bibitem{Martin2005}
M.~Haegele, T.~Skordas, S.~Sagert, R.~Bischoff, T.~Brog{\aa}rdh, and
  M.~Dresselhaus, ``White paper-industrial robot automation,''
  \url{http://www.euron.org/miscdocs/docs/euron2/year2/dr-14-1-industry.pdf},
  2005.

\bibitem{CARMEN2014}
J.~P. Buch, J.~S. Laursen, L.~C. S{\o}rensen, L.-P. Ellekilde, D.~Kraft, U.~P.
  Schultz, and H.~G. Petersen, ``Applying simulation and a domain-specific
  language for an adaptive action library,'' in \emph{Proceedings of
  International Conference on Simulation, Modeling, and Programming for
  Autonomous Robots (SIMPAR 2014)}, October 2014.

\bibitem{Angerer2013}
A.~Angerer, A.~Hoffmann, A.~Schierl, M.~Vistein, and W.~Reif, ``{Robotics API:
  Object-Oriented Software Development for Industrial Robots},'' \emph{Journal
  of Software Engineering of Robotics}, vol.~4, pp. 1--22, May 2013.

\bibitem{Thomas2013}
U.~Thomas, G.~Hirzinger, B.~Rumpe, C.~Schulze, and A.~Wortmann, ``A new skill
  based robot programming language using uml/p statecharts,'' in \emph{Robotics
  and Automation (ICRA), 2013 IEEE International Conference on}, May 2013, pp.
  461--466.

\bibitem{Simmons98}
R.~Simmons and D.~Apfelbaum, ``A task description language for robot control,''
  in \emph{in Proceedings of the Conference on Intelligent Robots and Systems
  (IROS}, 1998.

\bibitem{Henrik2010}
H.~M\"uhe, A.~Angerer, A.~Hoffmann, and W.~Reif, ``On reverse-engineering the
  kuka robot language,'' in \emph{1st International Workshop on Domain-Specific
  Languages and models for ROBotic systems (DSLRob’10)}, 2010.

\bibitem{Ellekilde2010}
L.-P. Ellekilde and J.~A. Jorgensen, ``Robwork: A flexible toolbox for robotics
  research and education,'' \emph{Robotics (ISR), 2010 41st International
  Symposium on and 2010 6th German Conference on Robotics (ROBOTIK)}, pp. 1
  --7, june 2010.

\bibitem{yokoyama2008principles}
T.~Yokoyama, H.~B. Axelsen, and R.~Gl{\"u}ck, ``Principles of a reversible
  programming language,'' in \emph{Proceedings of the 5th conference on
  Computing frontiers}.\hskip 1em plus 0.5em minus 0.4em\relax ACM, 2008, pp.
  43--54.

\end{thebibliography}



\end{document}